%% file: aaai_main.tex
\def\year{2020}\relax
\documentclass[letterpaper]{article} 
\usepackage{aaai20}  
\usepackage{times}  
\usepackage{helvet} 
\usepackage{courier}  
\usepackage[hyphens]{url}  
\usepackage{graphicx} 
\usepackage{graphicx}  
\usepackage{graphbox}
\input{math_commands.tex}

\usepackage{xcolor}

\renewcommand{\P}{P}
\newcommand{\Q}{Q}

\title{A Commentary on the Unsupervised Learning of Disentangled Representations}
\author{Francesco Locatello,\textsuperscript{\rm 2}\textsuperscript{\rm 3} Stefan Bauer,\textsuperscript{\rm 3} Mario Lucic,\textsuperscript{\rm 1} Gunnar R\"atsch,\textsuperscript{\rm 2} \\ \Large \textbf{Sylvain Gelly,\textsuperscript{\rm 1} Bernhard Sch\"olkopf,\textsuperscript{\rm 3} Olivier Bachem \textsuperscript{\rm 1}} \\ 
\textsuperscript{\rm 1}Google Research, Brain Team\\
\textsuperscript{\rm 2}ETH Zurich, Department for Computer Science\\
\textsuperscript{\rm 3}Max-Planck Institute for Intelligent Systems\\
Correspondence to francesco.locatello@inf.ethz.ch and
bachem@google.com
}
 
\begin{document}

\maketitle

\begin{abstract}
The goal of the \emph{unsupervised} learning of \emph{disentangled} representations is to separate the independent explanatory factors of variation in the data without access to supervision. 
In this paper, we summarize the results of~\cite{locatello2019challenging} and focus on their implications for practitioners. 
We discuss the theoretical result showing that the unsupervised learning of disentangled representations is fundamentally impossible without inductive biases and the practical challenges it entails. 
Finally, we comment on our experimental findings, highlighting the limitations of state-of-the-art approaches and directions for future research.
\end{abstract}

\section{Introduction}
In representation learning, we often have access to high-dimensional observations $\rvx$ (e.g., images or videos) without additional annotations. 
However, such observations are often assumed to be the manifestation of a set of low dimensional ground truth factors of variations $\rvz$. 
For example, the factors of variation in natural images may be pose, content, location of objects, and lighting conditions. 

The goal of representation learning is to learn a vector $r(\rvx)$ which is low dimensional and useful for any downstream task~\cite{bengio2013representation}. The key idea of disentangled representations is that they capture the information about the explanatory factors of variations independently: each factor of variation is separately represented in just a few dimensions of the representation~\cite{bengio2013representation}. In our example of natural images, we may wish to encode separately pose, content, location of objects, and lighting conditions.

\looseness=-1Disentangled representations hold the promise to be both interpretable, robust, and to simplify downstream prediction tasks~\cite{bengio2013representation}. Recently, disentanglement has been found useful for a variety of downstream tasks including fair machine learning~\cite{locatello2019fairness,creager2019flexibly}, abstract visual reasoning tasks~\cite{van2019disentangled} and real-world robotic data sets~\cite{gondal2019transfer}.

State-of-the-art approaches for unsupervised disentanglement learning are largely based on variants of \emph{Variational Autoencoders (VAEs)} \cite{kingma2013auto} where the encoder is further regularized to encourage disentanglement.

In this paper, we comment on some of the key contributions of~\cite{locatello2019challenging}:
\begin{itemize}
\item We discuss why it is impossible to learn disentangled representations for arbitrary data sets without supervision or inductive biases.
\item We provide a sober look at the performances of state-of-the-art approaches. We highlight challenges for model selection and identify critical areas for future research.
\item  To facilitate future research and reproducibility, we release a library to train and evaluate disentangled representations on standard benchmark data sets.
\end{itemize}
\section{Background}
\looseness=-1For the purpose of disentanglement learning, the world is modeled as a two-step generative process. First, we sample a latent variable $\rvz$ from a distribution with factorized density $p(\rvz)=\prod_{i=1}^dp(\rz_i)$. Each dimension of $\rvz$ corresponds to an independent factor of variation such as pose, content, locations of objects and lighting conditions in an image. Second, the observations are obtained as samples from $p(\rvx |\rvz)$.

The goal of disentanglement is to encode the factors of variation in a vector $r(\rvx)$ independently. The key idea is that a change in a dimension of $\rvz$ corresponds to a change in a dimension (or subset of dimensions) of $r(\rvx)$~\cite{bengio2013representation}. This definition has been further extended in the languages of group theory~\cite{higgins2018towards} and causality~\cite{suter2018interventional}.

\paragraph{Metrics} 
The lack of a formal definition of disentanglement resulted in a variety of different metrics. We assume access to $\rvz$ and characterize the structure of the statistical relations between $\rvz$ and $r(\rvx)$. Intuitively, we measure how the information about $\rvz$ is encoded in $r(\rvx)$.
The \emph{BetaVAE}~\cite{higgins2016beta} and \emph{FactorVAE}~\cite{kim2018disentangling} scores measures disentanglement by predicting the index of a fixed factor.
Other scores are typically composed of two steps: first, they estimate a matrix relating $\rvz$ and $r(\rvx)$. The \emph{Mutual Information Gap (MIG)}~\cite{chen2018isolating} and \emph{Modularity}~\cite{ridgeway2018learning} estimate the pairwise mutual information matrix, \emph{DCI Disentanglement}~\cite{eastwood2018framework} the feature importance predicting $\rvz$ from $r(\rvx)$ and the \emph{SAPscore}~\cite{kumar2017variational} the predictability of $\rvz$ from $r(\rvx)$. 
Second, this matrix is aggregated to obtain a score by computing some normalized gap either row- or column-wise. For more details, see Appendix C of~\cite{locatello2019challenging}.

\paragraph{Methods} 
\looseness=-1In \emph{Variational Autoencoders (VAEs)} \cite{kingma2013auto}, one assumes a prior $P(\rvz)$ on the latent space and parameterizes the conditional probability $\P(\rvx|\rvz)$ using a deep neural network (i.e., a \textit{decoder network}). The posterior distribution is approximated by a variational distribution $\Q(\rvz|\rvx)$, again parameterized using a deep neural network (i.e., an \textit{encoder network}). The model is then trained by maximizing a variational lower-bound to the log-likelihood and the representation $r(\rvx)$ is usually taken to be the mean of the encoder distribution. To learn disentangled representations, state-of-the-art approaches enrich the VAE objectives with a suitable regularizer.

The $\beta$-VAE~\cite{higgins2016beta} and AnnealedVAE~\cite{burgess2018understanding} constrain the capacity of the VAE bottleneck. The intuition is that recovering the factors of variation is the most efficient compression scheme to achieve good reconstruction~\cite{PetJanSch17}. The Factor-VAE~\cite{kim2018disentangling} and $\beta$-TCVAE both penalize the total correlation of the aggregated posterior $Q(\rvz) = \int Q(\rvz|\rvx) d\rvx$ (i.e. the encoder distribution after marginalizing the training data). The DIP-VAE variants~\cite{kumar2017variational} match the moments of the aggregated posterior and a ``disentanglement prior'', which in practice is simply a factorized distribution. We refer to Appendix B of \cite{locatello2019challenging} for a more detailed description.

\section{Theoretical impossibility}
Theorem~1 in~\cite{locatello2019challenging} states that the unsupervised learning of disentangled representation is impossible for arbitrary data sets. Even in the infinite data regime, where supervised learning algorithms such as k-nearest neighbours classifiers are consistent, no model can find a disentangled representation observing samples from $P(\rvx)$ only. This theoretical result motivates the need for either implicit supervision, explicit supervision, or inductive biases. 

The key idea is that we can construct two generative models whose latent variables $\rvz$ and $f(\rvz)$ are entangled with each other but have the same marginal distribution over $\rvx$, i.e., the same $P(\rvx)$. If a representation is disentangled with one of these generative models it must be entangled with the other by construction. Observing only samples from $P(\rvx)$, it is impossible to distinguish which model $r(\rvx)$ should disentangle: both $\rvz$ and $f(\rvz)$ are equally plausible and ``look the same'' as they produce the same $\rvx$ with the same probability.

Note that Theorem~1 in~\cite{locatello2019challenging} does not account for the structure that real world generative models may exhibit. Inductive biases on both the models and the data may be sufficient to learn disentangled representations in practice as certain solutions may be favored instead of others, i.e., some model may naturally converge to a solution that disentangles the true $\rvz$ instead of $f(\rvz)$. 
Similar results have been obtained in the context of non-linear ICA~\cite{hyvarinen1999nonlinear} where i.i.d. data is known to be insufficient for identifiability, in general. 
\paragraph{Implications} We proved that the unsupervised learning of disentangled representations is in general impossible without inductive biases on both methods and data sets. We argue that future work should make the role of inductive biases or supervision more explicit.

\section{Disentanglement in practice}
In this section, we highlight the implications of some of the empirical results of~\cite{locatello2019challenging}. We implemented six recent unsupervised disentanglement learning methods as well as six disentanglement metrics from scratch. Overall, we trained over $12000$ models and computed over $150000$ scores on seven data sets and 50 random seeds.\footnote{Reproducing our results requires approximately 2.52 GPU years (NVIDIA P100).} We refer to Section 5 of~\cite{locatello2019challenging} for more details and a richer quantitative description.

\begin{figure*}
 \centering\includegraphics[align=c, width=0.25\textwidth]{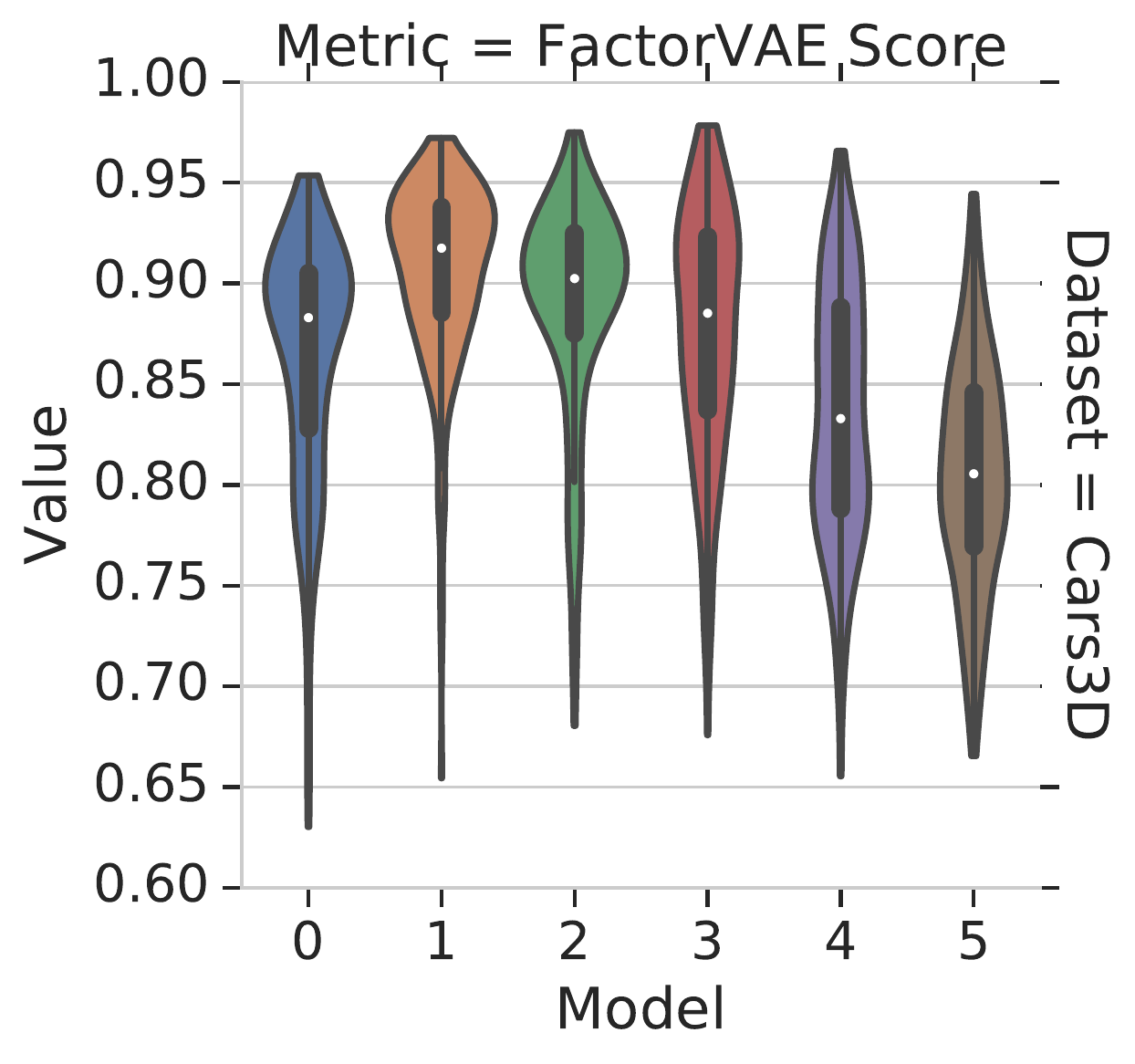}%
 \centering\includegraphics[align=c, width=0.3\textwidth]{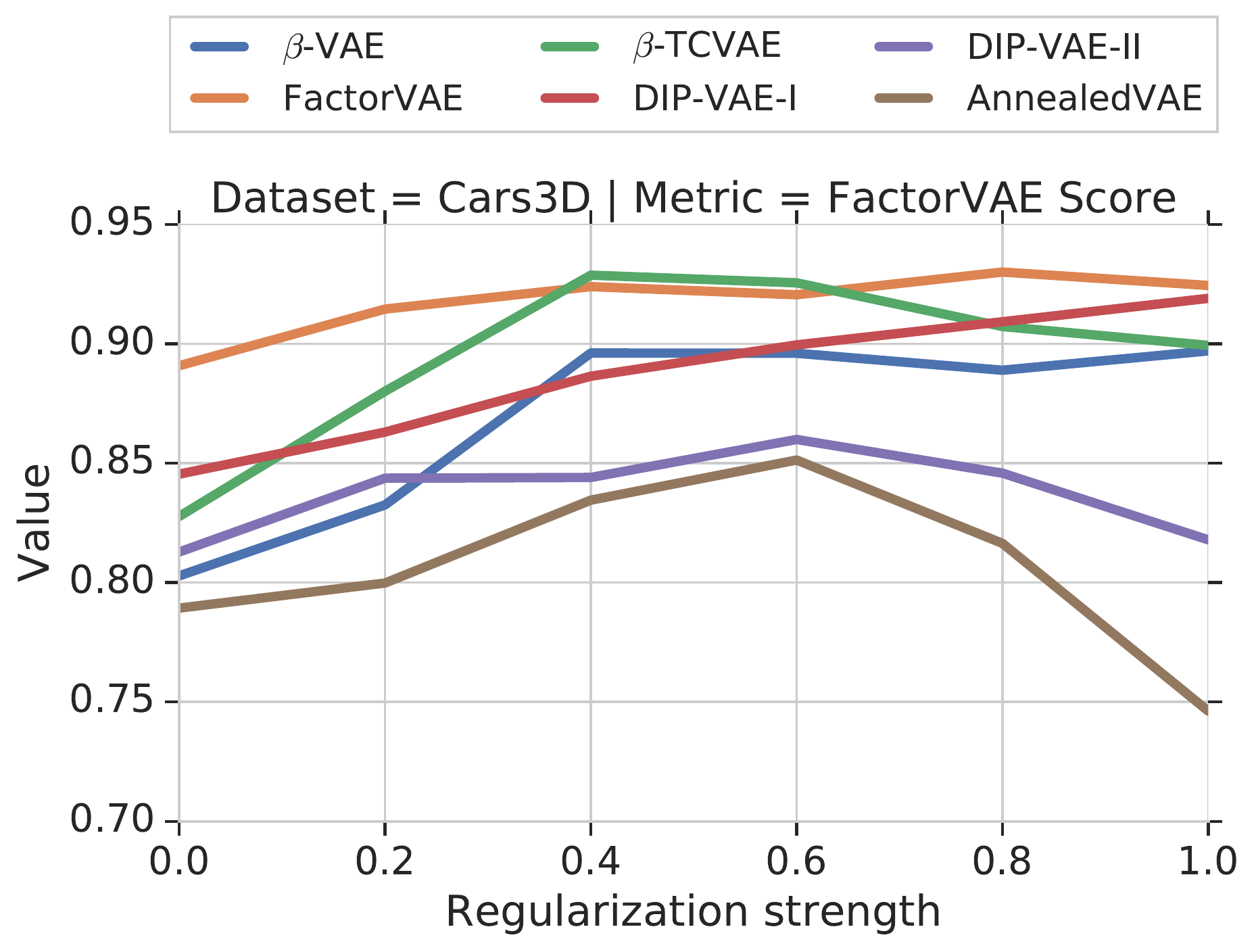}%
\includegraphics[align=c, width=0.32\textwidth]{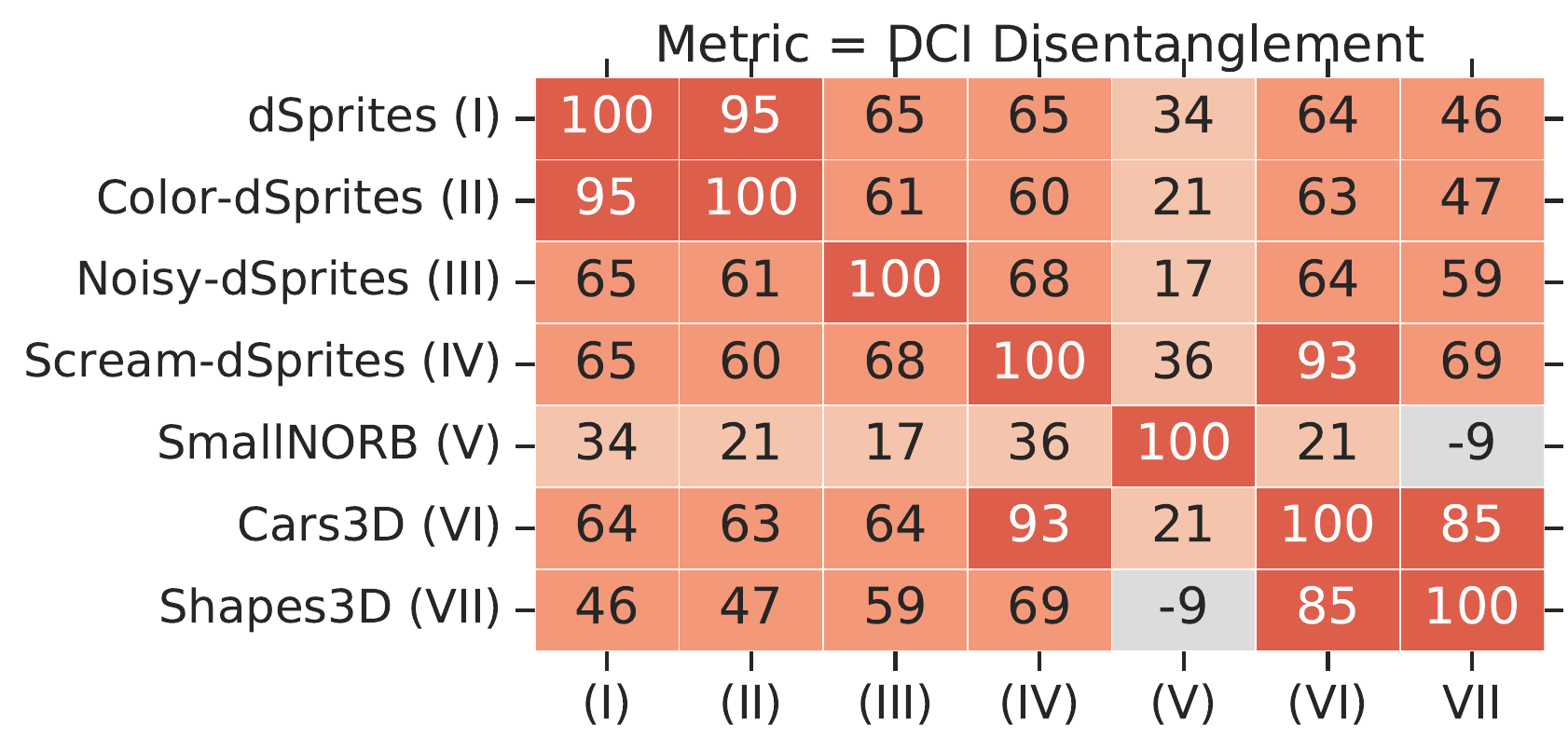}
\caption{
(left) FactorVAE score for each method on Cars3D. Models are abbreviated: 0=$\beta$-VAE, 1=FactorVAE, 2=$\beta$-TCVAE, 3=DIP-VAE-I, 4=DIP-VAE-II, 5=AnnealedVAE. The variance is due to different hyperparameters and random seeds. We observe that the scores are heavily overlapping. (center) FactorVAE score vs hyperparameters for each score on Cars3d. There seems to be no model dominating all the others and for each model there does not seem to be a consistent strategy in choosing the regularization strength. (right) Rank-correlation of DCI disentanglement metric across different data sets. Good hyperparameters seem to transfer especially between dSprites
and Color-dSprites. }\label{figure:main_figure}
\end{figure*}

\subsection{Which method should be used?}
This first question is particularly relevant for practitioners interested in the benefits of disentanglement methods off-the-shelf. In Figure~\ref{figure:main_figure} (left), e observe that the choice of the objective function seems to matter less than the choice of hyperparameters and seed. In particular, only $37\%$ of the observed variance in the models can be explained by the choice of the objective function.  Since our trained models exhibit such a large variance, it appears to be crucial to identify good hyperparameters and runs.

\paragraph{Implications} 
It is not clear which method should be used and choosing good hyperparameters, and selecting good runs seem to be matter more.

\subsection{How to choose the hyperparameters?}
We investigated whether we may find “rules of thumb” for selecting good hyperparameters. In Figure~\ref{figure:main_figure} (center), we plot the median FactorVAE score for different regularization strengths for each method on Cars3D. We observe that no method is consistently better than all the others and there does not seem to be an obvious trend that can be used to maximize disentanglement scores. In Figure~\ref{figure:main_figure} (right), we test whether good hyperparameter settings may be transferred across data sets. We observe that at the distribution level there appears to be some correlation between the disentanglement scores across the different data sets.

\paragraph{Implications} 
There is no clear rule of thumb, but transfer across data sets may help. Note that we still cannot distinguish between a good and a bad training run.

\subsection{How to select the best model from a set of trained models?}
\begin{figure*}
 \centering\includegraphics[align=c, width=0.2\textwidth]{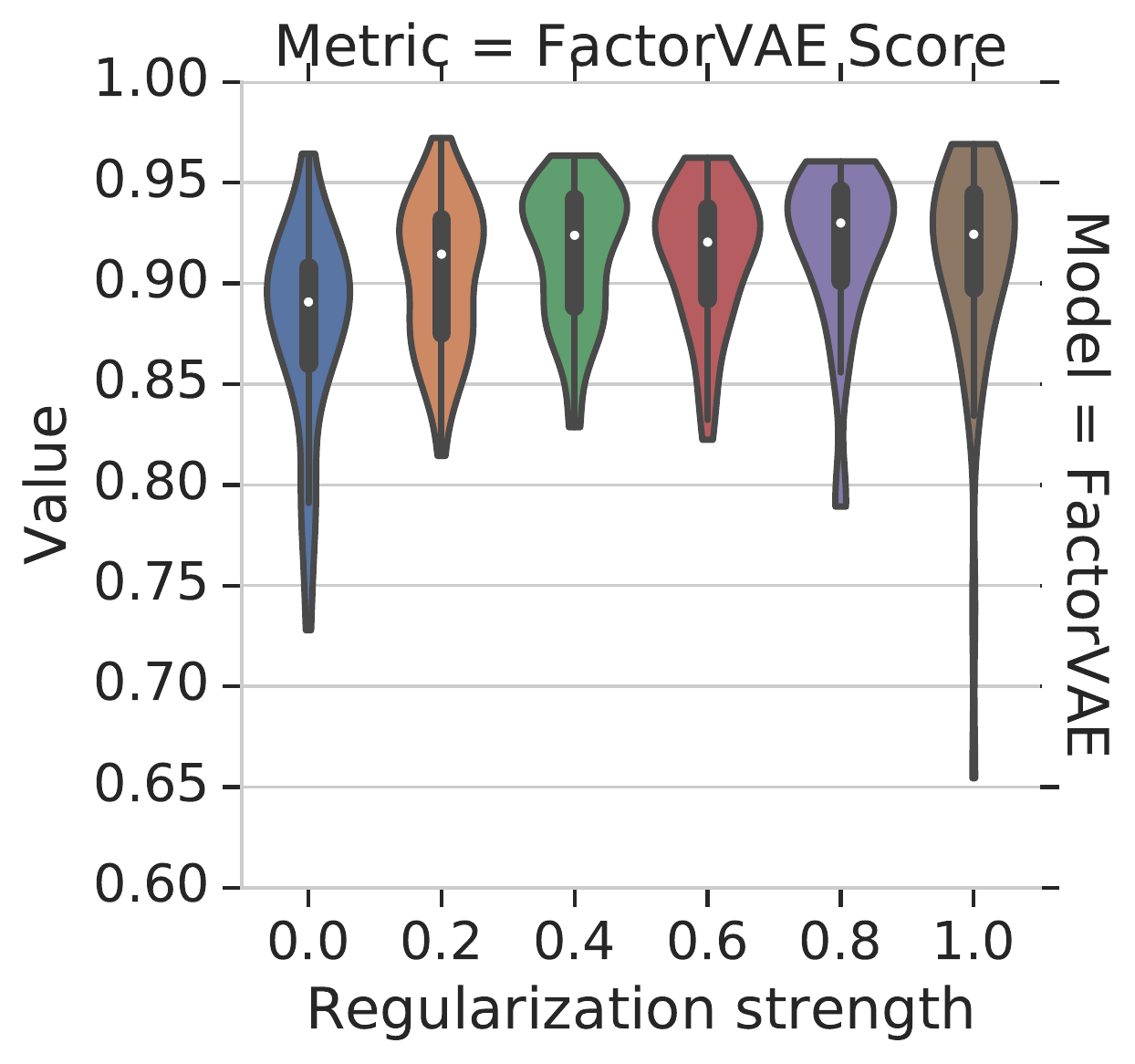}
 \centering\includegraphics[align=c, width=0.3\textwidth]{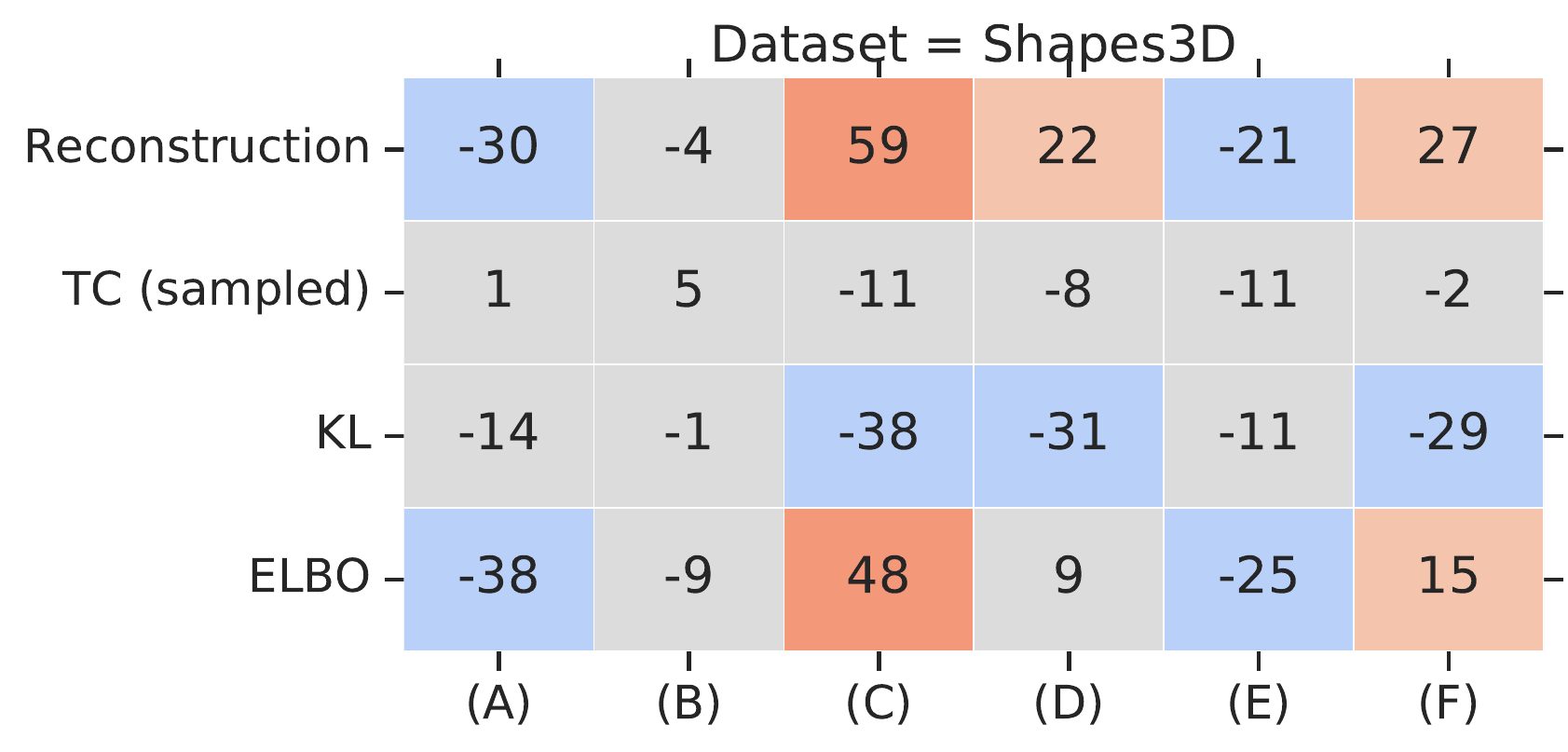}
\caption{
(left) Distribution of FactorVAE scores for FactorVAE model for different regularization strengths on Cars3D. (right) Rank correlation between unsupervised scores and disentanglement metrics on Shapes3D. Metrics are abbreviated: (A)=BetaVAE Score, (B)=FactorVAE Score, (C)=MIG , (D)=DCI
Disentanglement, (E)=Modularity, (F)=SAP.}\label{figure:second_figure}
\end{figure*}

\looseness=-1First, we note that the transfer of hyperparmeters does not reliably outperforms random model selection: it improves only $59.3\%$ of the times. To understand why this is the case we plot in Figure~\ref{figure:second_figure} (left) the distribution of Factor VAE models evaluated with the FactorVAE score on Cars3D. We observe that randomness has a substantial impact on the representation as a good run with bad hyperparameters can easily outperform a bad run with the best hyperparameters. Finally, we check whether the unsupervised training metrics may be used for model selection.  In Figure~\ref{figure:second_figure} (right), we observe that the training metrics appear to be rather uncorrelated with disentanglement.

\paragraph{Implications} Unsupervised model selection remains an open research challenge. Transfer of good hyperparameters does not seem to work and we did not find a way to distinguish between good and bad runs without supervision.

\section{Directions of future research}
Finally, we discuss the critical open challenges in disentanglement and some of the lessons we learned with this study.

\paragraph{Inductive biases and implicit and explicit supervision.}
Our results highlights an overall need for supervision. In theory, inductive biases are crucial to distinguish among equally plausible generative models. In practice we did not find a reliable strategy to choose hyperparameters without supervision. Recent work~\cite{duan2019heuristic} proposed a stability based heuristic for unsupervised model selection. Further exploring these techniques may help us understand the practical role of inductive biases and implicit supervision. Otherwise, we advocate to consider different settings, for example when limited explicit~\cite{locatello2019disentangling} or weak supervision~\cite{bouchacourt2017multi,gresele2019incomplete} is available.

\paragraph{Experimental setup and diversity of data sets.}
\looseness=-1Our study highlights the need for a sound, robust, and reproducible experimental setup on a diverse set of data sets.
In our experiments, we observed that the results may be easily misinterpreted if one only looks at a subset of the data sets. As current research is typically focused on the synthetic data sets of~\cite{higgins2016beta,reed2015deep,lecun2004learning,kim2018disentangling,locatello2019challenging} --- with only a few recent exceptions~\cite{gondal2019transfer} --- we advocate for insights that generalize across data sets rather than individual absolute performance.
For this reason, we released \texttt{disentanglement\_lib}\footnote{\url{https://github.com/google-research/disentanglement_lib}}, a library to facilitate reproducible research on disentanglement. Our library allows to train and evaluate state-of-the-art disentangled representations on common benchmark data sets and produces automatic visualizations for visual inspection on all the trained models. Furthermore, we released over $10000$ trained models, which can be used as baselines for future research.

\section{Acknowledgements}
The authors thank Ilya Tolstikhin, Paul Rubenstein and Josip Djolonga for helpful discussions and comments. This research was partially supported by the Max Planck ETH
Center for Learning Systems, by an ETH core grant
(to Gunnar R\"atsch) and a Google Ph.D. Fellowship to FL. This work was partially done while
FL was at Google Research Zurich.
\bibliographystyle{aaai}
\bibliography{main}

\end{document}

%% file: math_commands.tex
\usepackage{amsmath,amsfonts,bm}

\def\eqref#1{equation~\ref{#1}}

\def\1{\bm{1}}

\def\rz{{\textnormal{z}}}

\def\rvx{{\mathbf{x}}}

\def\rvz{{\mathbf{z}}}

\DeclareMathAlphabet{\mathsfit}{\encodingdefault}{\sfdefault}{m}{sl}
\SetMathAlphabet{\mathsfit}{bold}{\encodingdefault}{\sfdefault}{bx}{n}

